\documentclass{article}

\usepackage{PRIMEarxiv}

\usepackage[utf8]{inputenc} 
\usepackage[T1]{fontenc}    
\usepackage[english]{babel} 
\usepackage{hyperref}       
\usepackage{url}            
\usepackage{booktabs}       
\usepackage{amsfonts}       
\usepackage{nicefrac}       
\usepackage{microtype}      
\usepackage{lipsum}         
\usepackage{fancyhdr}       
\usepackage{graphicx}       
\graphicspath{{images/}}    

\usepackage[compress]{cite}
\usepackage{textcomp}
\usepackage[table]{xcolor}
\usepackage{subcaption}     
\usepackage{array,multirow}
\usepackage{amsmath,amssymb,mathrsfs}
\usepackage[font=small,labelfont=bf]{caption}
\usepackage{algorithm}
\usepackage{algorithmicx}
\usepackage{algpseudocode}
\usepackage{bbm}
\usepackage{bm}
\usepackage{adjustbox}
\usepackage{siunitx}        
\usepackage{lmodern}
\usepackage{fancyhdr}

\title{Memory- and Latency-Constrained Inference of Large Language Models via Adaptive Split Computing}

\author{
    Mingyu~Sung,
    Vikas~Palakonda,
    Suhwan~Im,
    Sunghwan~Moon, \\
    Il-Min~Kim,
    Sangseok~Yun,
    and~Jae-Mo~Kang
    \thanks{
        Mingyu Sung, Vikas~Palakonda, Suhwan~Im and Jae-Mo Kang are with the Department of Artificial Intelligence, Kyungpook National University, Daegu, South Korea (Corresponding author: Jae-Mo Kang, e-mail: jmkang@knu.ac.kr).
    }
    \thanks{
        Il-Min Kim is with the Department of Electrical and Computer Engineering, Queen's University, Kingston, K7L 3N6, Canada.
    }%
    \thanks{
        Sangseok Yun is with the Department of Information and Communications Engineering, Pukyong National University, Busan 48513, South Korea (Corresponding author: Sangseok Yun, e-mail: ssyun@pknu.ac.kr).
    }
}

\begin{document}
\maketitle

\begin{abstract}
Large language models (LLMs) have achieved near-human performance across diverse reasoning tasks, yet their deployment on resource-constrained Internet-of-Things (IoT) devices remains impractical due to massive parameter footprints and memory-intensive autoregressive decoding. While split computing offers a promising solution by partitioning model execution between edge devices and cloud servers, existing approaches fail to address the unique challenges of autoregressive inference, particularly the iterative token generation process and expanding key-value (KV) cache requirements.
This work introduces the first autoregressive-aware split computing framework designed explicitly for LLM deployment on edge devices. Our approach makes three key contributions. First, we develop one-point split compression (OPSC), a mixed-precision quantization scheme that prevents out-of-memory failures by strategically partitioning models into front-end and back-end segments with different precision levels. Second, we propose a two-stage intermediate compression pipeline that combines threshold splitting (TS) and token-wise adaptive bit quantization (TAB-Q) to preserve accuracy-critical activations while dramatically reducing communication overhead. Third, we formulate a unified optimization framework that jointly selects optimal split points, quantization settings, and sequence lengths to satisfy strict memory and latency constraints.
Extensive evaluations across diverse LLMs and hardware platforms demonstrate superior performance compared to state-of-the-art quantization methods, including SmoothQuant, OmniQuant, and Atom. The framework achieves a 1.49× inference speedup and significant communication overhead reduction while maintaining or improving model accuracy. Notably, the approach enables deployment of models with hundreds of gigabytes of memory requirements on edge devices with severely constrained resources, making large-scale LLMs practically accessible for real-time IoT applications.
\end{abstract}


\section{INTRODUCTION}
Deep neural networks (DNNs) have revolutionized fields such as computer vision, mobile sensing, and the Internet of Things (IoT), driving significant advancements in data analysis and decision-making. The groundbreaking transformer architecture introduced by Vaswani \textit{et al.}~\cite{vaswani2017attention} has further accelerated progress in natural language processing (NLP), facilitating the development of sophisticated large language models (LLMs) \cite{devlin2018bert,NEURIPS2020_1457c0d6,dubey2024llama}. These models demonstrate remarkable capabilities in tasks including natural language understanding, natural language generation, complex reasoning, and code generation \cite{minaee2024large,patwardhan2023transformers}. They power various groundbreaking applications, such as ChatGPT, GitHub Copilot, and the new Bing search experience \cite{zhou2024survey}. Beyond NLP, LLMs have also shown promise in super-resolution, IoT sensor processing, image generation/synthesis, and voice processing \cite{liu2023summary}.

Despite their exceptional performance, LLMs impose substantial computational and energy costs during both training and inference \cite{zhao2023survey}. This challenge has become a critical operational burden for large technology corporations deploying LLMs as cloud-based services, particularly as user demand escalates exponentially.

Modern edge devices possess considerable processing capabilities and can handle substantial portions of these computations. However, the prevailing server-centric operational model concentrates the entire computational and financial burden on centralized infrastructure. This paradigm results in severe underutilization of high-performance edge resources, while demand for real-time, on-device processing in applications such as personalized AI continues to surge \cite{wang2021comprehensive}.. 

\subsection{Related Work and Motivations}
\subsubsection{Merits of Split Computing}
To bridge this gap, split computing (SC) has emerged as a promising paradigm, enabling collaborative inference that leverages the capabilities of both edge devices and cloud servers \cite{kang2017neurosurgeon}. In this model, an edge device executes the initial network layers and offloads the intermediate outputs to the server, which completes the inference task. This method presents a viable alternative by distributing the computational workload, thus alleviating the server's burden while utilizing the processing power of edge devices. However, its effectiveness depends on identifying optimal split points that balance on-device computation against communication overhead.
Studies have explored various strategies to reduce inference latency, including selective splitting, architectural optimization, and intermediate feature compression \cite{sung2025decomesc,yun2022cooperative,bakhtiarnia2023dynamic,cohen2020lightweight,oh2023communication}. However, most of these efforts focus on image-based models, leaving SC for LLMs remains underexplored.

\vspace{-0.5cm}
\subsubsection{Limitations of Split Computing for LLMs}
Deploying LLMs on edge devices presents two crucial challenges: high computational demand and stringent memory constraints\footnote{For instance, GPT-3 requires about 1.7 s to process a 512-token input and generate a 32-token output on eight Nvidia A100 GPUs \cite{lin2023pushing}, which is impractical for most edge devices without extensive optimization.}.
While SC should split a DNN between an edge device and the cloud, modern LLMs behave \emph{autoregressively}, generating text by iteratively refeeding newly produced tokens through the entire model.
This property complicates conventional SC in two ways. First, each freshly generated token must pass through all layers again, undermining naive strategies that place only the first few layers on the edge. Second, the repeated transformations of the growing token sequence can easily lead to out-of-memory (OOM) problems if the edge device lacks sufficient memory for the expanding intermediate states.

Although recent works on SC \cite{bajpai2023splitee,ohta2023lambda} have addressed smaller transformer models or partially transmitted hidden layer outputs, no study has fully tackled the massive parameter sizes and iterative nature of large-scale LLMs. For instance, Bajpai \textit{et~al.}~\cite{bajpai2023splitee} employed confidence-based splitting; their evaluation was limited to RoBERTa (123M parameters) without considering the complexities introduced by autoregressive inference. Ohta \textit{et~al.}~\cite{ohta2023lambda} focused on privacy-sensitive intermediate outputs but did not explore split-layer optimization or the increased memory footprint from token-by-token generation. 
Consequently, naive application of existing SC frameworks to LLMs may yield suboptimal performance, OOM failures, or underutilization of edge resources.

\subsubsection{Need for Split Computing in Deploying LLMs}

\begin{figure}[htbp]
  \centering
  \includegraphics[width=0.5\textwidth]{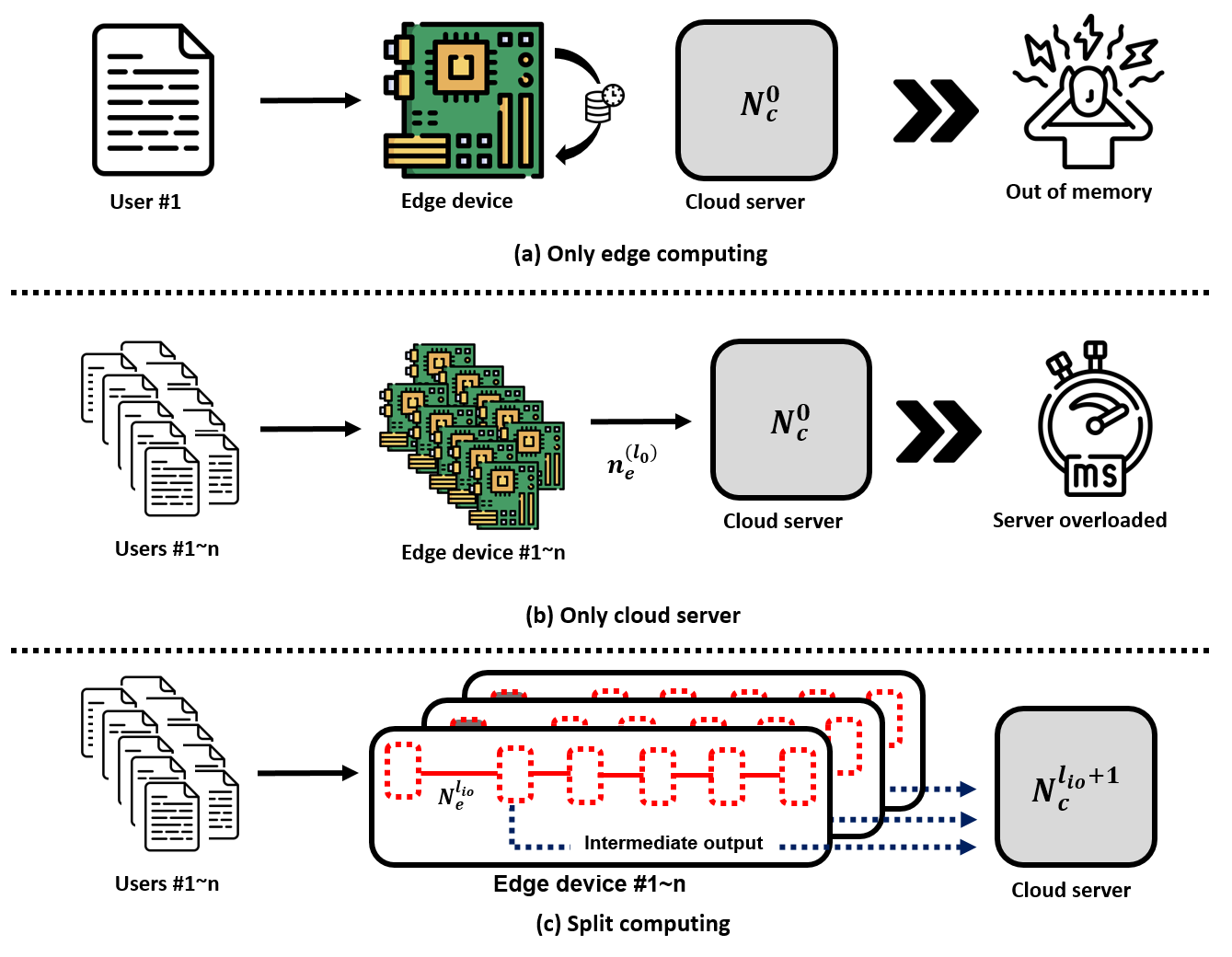}
  \caption{Schematic diagram of three scenarios for deploying LLM to edge devices. (a) local computing, (b) edge computing, and (c) split computing (SC)}
  \label{Need}
\end{figure}

Fig.~\ref{Need} highlights the need for SC in deploying LLMs by illustrating three typical deployment scenarios: (a) local computing  (b) edge computing, and (c) SC \cite{matsubara2022split}. In the local computing scenario, researchers have attempted to compress and quantize LLMs to fit on a single edge device \cite{yi2023edgemoe,xiao2023smoothquant, shao2023omniquant, zhao2024atom,lin2023pushing,qin2023enabling}, but these methods encounter significant memory limitations and performance degradation. In contrast, the edge computing scenario offloads all inference tasks to a cloud server, with edge devices functioning solely as data transmitters. Although this simplifies on-device requirements, it risks overloading the server and underutilizing edge capabilities. By comparison, the SC approach enables LLM deployment by distributing computational and memory burdens between edge devices and the cloud, making it ideal for resource-constrained environments.

\subsection{Contributions}
Given these challenges, we propose an SC strategy centered on three core objectives for efficiently deploying LLMs:

\begin{enumerate}
    \item \textbf{Accommodating Autoregressive Inference:}
    Because LLMs re-invoke their entire architecture for each newly produced token, an SC design must effectively handle iterative data flow and avoid repeated large-scale data transfers between the edge and the cloud. In particular, we focus on leveraging the key–value (KV) cache.
    \item \textbf{Compressing Massive Parameters:}
    Modern LLMs contain billions of parameters; thus, advanced compression and mixed-precision techniques are essential for preventing OOM errors and excessive latency on edge devices.
    \item \textbf{Maximizing Edge Utilization:}
    Offloading most computations to the cloud wastes the increasing capabilities of edge hardware and risks bottlenecks in cloud resources. By intelligently selecting split layers and employing local inference, one can avoid overburdening the cloud and leverage the computational potential of the edge device.
\end{enumerate}
By jointly addressing these three aspects, the proposed SC framework preserves high accuracy and minimizes resource usage, even under the demanding conditions of autoregressive generation. This approach ensures that large-scale LLMs become more accessible for diverse IoT and real-time AI scenarios, balancing on-device performance and cloud-based scalability.

The main contributions of this work are summarized as follows:
    
    

\begin{enumerate}
   \item \textbf{Autoregressive-Aware Split Computing Framework:} This work proposes an autoregressive-aware split computing framework explicitly designed for large language model (LLM) inference. The framework systematically addresses token-by-token generation and key-value (KV) cache expansion through one-point split compression (OPSC), a mixed-precision quantization scheme that prevents out-of-memory failures on edge devices.
   
   \item \textbf{Two-Stage Intermediate Feature Compression with Unified Optimization:} A novel two-stage compression pipeline combines threshold splitting (TS) and token-wise adaptive bit quantization (TAB-Q) to minimize communication overhead while preserving accuracy-critical activations. Additionally, a unified optimization framework jointly determines split points, quantization configurations, and sequence lengths under strict memory and latency constraints.
   
   \item \textbf{Comprehensive Validation Across Models and Platforms:} Extensive experiments across diverse LLM architectures and hardware platforms demonstrate the effectiveness and generalizability of the proposed framework, establishing its practicality for resource-constrained edge-cloud deployments.
\end{enumerate}

The remainder of the paper is organized as follows. Section II introduces the framework of the proposed method, and Section III presents the experimental setup and result analysis. Finally, Section IV concludes the paper.

\begin{table}
\caption{List of notations used in this paper.}
\centering
\renewcommand{\arraystretch}{1.2} 
\begin{tabular}{m{2.6cm}|m{5.4cm}}
\hline 
\hline
\textbf{Notation} & \textbf{Description} \\
\hline
$D$ & Dimension of each head \\
\hline
$H$ & Number of attention heads \\
\hline
$T_w$ & Current hidden state tensor with length $w$ \\
\hline
$\bar{W}$ & Maximum token sequence length \\
\hline
$Q^w = \{Q^w_1, Q^w_2\}$ & Set of front/back weight quantization bits \\
\hline
$Q^a = \{Q^a_1, Q^a_2\}$ & Set of front/back activation quantization bits \\
\hline
$M, D$ & Memory and delay constraint of edge device \\
\hline
$M(\ell_w, Q^w)$ & Total memory footprint of OPSC \\
\hline
$B_{kv}(w, \ell; Q^a)$ & Incremental memory usage for KV cache \\
\hline
$B_{io}$ & Intermediate output size with all parameters \\
\hline
$T_{above}, T_{below}$ & Split tensors above/below threshold \\
\hline
$\tau$ & Threshold for splitting \\
\hline
$\Delta$ & Distortion tolerance for TAB-Q \\
\hline
$A_\Delta$ & Acceptable accuracy drop \\
\hline
$\bar{Q}_a$ & Maximum activation quantization bits \\
\hline
$\gamma$ & Signal-to-noise ratio \\
\hline
$\Psi(Q^a)$ & Total activation-bit precision measure \\
\hline
$\epsilon$ & Target outage probability \\
\hline
$P_o(R)$ & Outage probability at rate R \\
\hline
$L_\epsilon(D_{tx}; R)$ & Worst-case latency for data size $D_{tx}$ \\
\hline
$L_t(T_w, \ell, Q^a, I_{kv}; R)$ & Total edge-device inference latency \\
\hline
$I_{kv}$ & Indicator function for KV cache inclusion \\
\hline
$g(R)$ & Rate optimization function \\
\hline
\end{tabular}
\end{table}

\section{Proposed Framework}
\subsection{One-Point Split Compression for Memory Constraint}
This work considers a scenario in which multiple memory-constrained edge devices perform inference tasks for an LLM via SC with a single cloud server (Fig.~\ref{Need}(c)). The primary goal of the proposed scheme is to maximize edge device utilization while satisfying memory and latency constraints.

Strict edge-side memory budgets inherently cap the deployable model size, rendering aggressive LLM compression indispensable. Consequently, a number of techniques have been proposed (e.g., \cite{xiao2023smoothquant, shao2023omniquant, zhao2024atom}). These lightweight approaches, however, introduce an accuracy–compression trade-off: excessive compression degrades performance, whereas insufficient compression can still trigger out-of-memory failures under SC deployments.

\begin{figure}[htbp]
  \centering
  \includegraphics[width=0.5\textwidth]{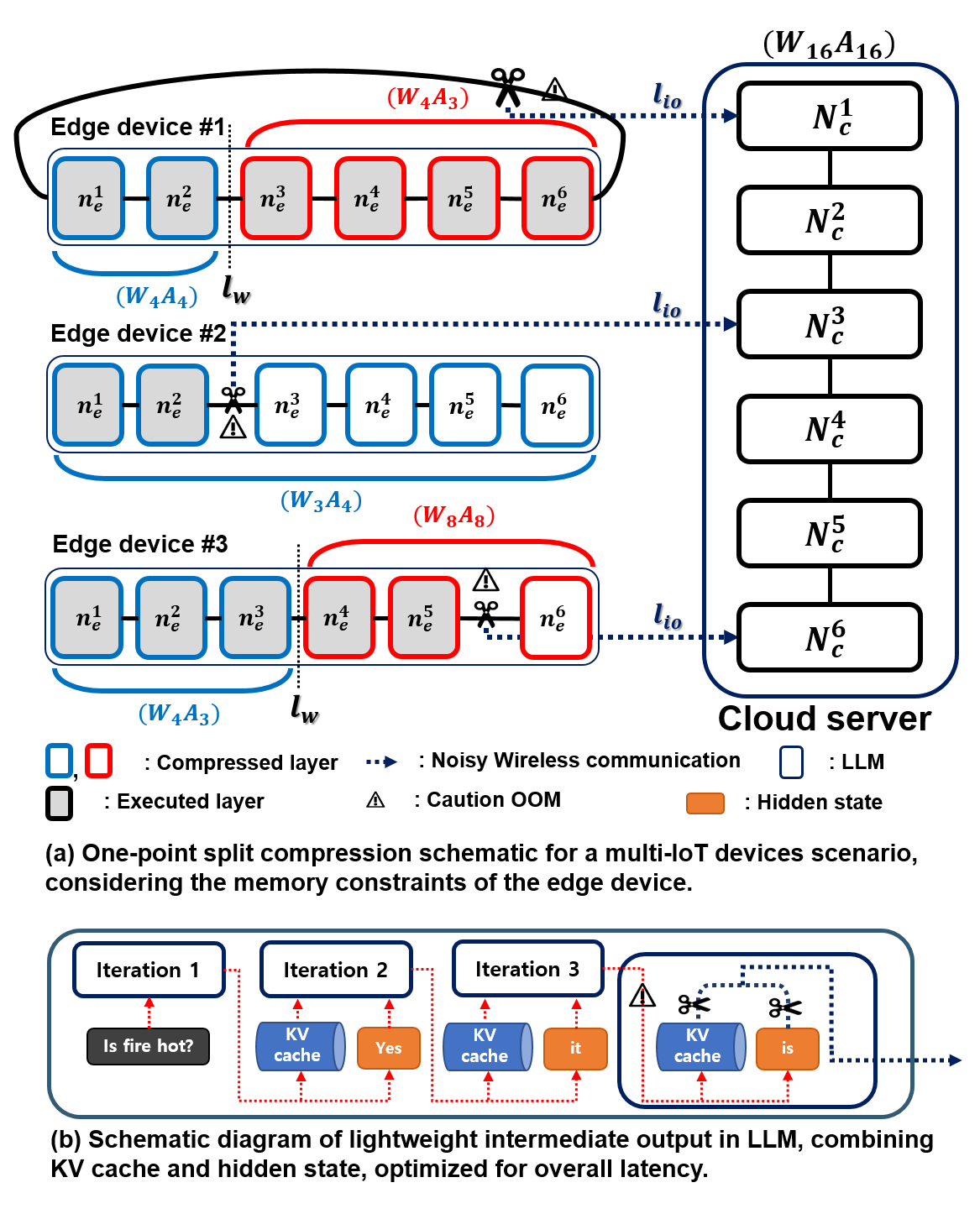}
  \caption{(a) One-point split compression schematic. (b) Intermediate output of LLM}
  \label{system_model}
\end{figure}

To address this issue, we propose a \emph{one-point split compression}~(OPSC) method. It is particularly beneficial for the cloud server to maintain only a single, high-precision model\footnote{Compressing weights or biases of a DNN causes information loss. Thus, the original precision of the layers must be preserved to minimize any degradation in inference accuracy \cite{sung2025decomesc}.}.
As illustrated in Fig.~\ref{system_model}(a), OPSC employs a mixed-precision technique\footnote{Mixed-precision techniques are widely used for lightweight LLMs (e.g., \cite{jiang2024mixtral,frantar2022gptq,coenders2023accounting,lin2024awq,xiao2023smoothquant,zhao2024atom}) and achieve excellent trade-offs between compression ratio and performance.} to apply various quantization levels at a single split point in the model. In contrast to existing methods that assign layerwise quantization, thereby requiring separate fine-tuning for each device,
OPSC partitions the model into front and back segments, each quantized differently, leading to less distortion than layerwise quantization and eliminating the need for additional fine-tuning for every device configuration. 
Consequently, it accommodates diverse memory constraints and compression settings across edge devices while preserving model accuracy, making it suitable for distributed environments with heterogeneous hardware resources\footnote{We adopt the state-of-the-art LLM compression framework, Atom \cite{zhao2024atom}, which considerably reduces memory usage and computational cost while maintaining high accuracy.}. Note that OPSC uses mixed-precision quantization to optimize model weights and parameters on edge devices.

Let us define the memory footprint of OPSC as follows:
\begin{align}
M\bigl(\ell_w, \mathcal{Q}^w\bigr)
&=\sum_{i=1}^{\ell_{w}} B_w\bigl(i; Q_{w1}\bigr) 
\;+\; \sum_{j=\ell_{w}+1}^{L} B_w\bigl(j; Q_{w2}\bigr),
\label{eq:memorysplit}
\end{align}
where $M\bigl(\ell_w, \mathcal{Q}^w\bigr)$ is the total memory footprint of the front (layers $1$ to $\ell_w$) and back segments (layers $\ell_w+1$ to $L$) under distinct quantization levels. Here, $\mathcal{Q}^w = \{Q_{w1}, Q_{w2}\}$ denotes the front/back weight quantization bits.

\subsection{Intermediate Output of LLM}
\subsubsection{Intermediate Output Definition}
The core component of an LLM is its transformer decoder block, incorporating multi-head attention (MHA) to focus on salient information via query (Q), key (K), and value (V) vectors.
In an \emph{autoregressive} generation scheme, the model produces each token sequentially; once a token is generated, it is fed back into the model from the first layer. As the sequence length grows to \(w\), the dimensions of \(Q\), \(K\), and \(V\) update accordingly:
\[
Q, K, V \in \mathbb{R}^{w \times HD} 
\;\;\longrightarrow\;\; 
Q', K', V' \in \mathbb{R}^{(w+1) \times HD}.
\]
However, repeatedly reprocessing all tokens is computationally expensive. To alleviate this, many LLMs use a \emph{KV cache} to store key and value states for previously generated tokens,\footnote{An LLM processes tokens one by one, appending the key and value for each newly generated token to the KV cache. This allows efficient “look back” at previously processed tokens without rerunning all computations.} enabling \emph{incremental decoding} in which only the newly generated token is processed at each step.

Although KV caching significantly improves inference speed, it may cause OOM errors on memory-constrained devices \cite{kwon2023efficient}. Hence, our SC framework carefully monitors the growth of the KV cache. We quantify how activation bit-widths affect KV-cache memory at each layer as follows:
\begin{equation}
\label{eq:bkv_lth_stage}
\begin{aligned}
B_{kv}\bigl(w, \ell; \mathcal{Q}^a\bigr)
  &= 2 \sum_{k=1}^{\ell} \Bigl(T_{w}\, Q_{a,k}\Bigr)
     + 2 \sum_{k=\ell+1}^{L} \Bigl(T_{w-1}\, Q_{a,k}\Bigr) \\
  &\quad + \underbrace{HD\, Q_{a,\ell}}_{\substack{\text{Memory usage of}\\\text{$w$-th token at layer $\ell$}}}, T_{w}\in \mathbb{R}^{w \times HD}.
\end{aligned}
\end{equation}
With \(T_{w}=wHD\) representing the size of the key (or value) tensor for \(w\) tokens, the required edge memory is
\[
Q_{a,k} =
\begin{cases}
Q_{a1}, & k < \ell_{w},\\
Q_{a2}, & k \ge \ell_{w},
\end{cases}
\qquad
\ell_{w}\!:\ \text{Eq.~\eqref{eq:memorysplit}}.
\]
The first term of Eq.~\eqref{eq:bkv_lth_stage} captures the KV tensors of the newly generated token \(w\) for layers computed on the edge (\(1\le k\le\ell\)).  
The second term accounts for the KV tensors of the \((w-1)\) previously generated tokens, which must still be buffered for the \(\ell+1\le k\le L\) layers executed in the cloud.  
Finally, the \(HD\,Q_{a,\ell}\) term represents the transient hidden state of token \(w\) at layer~\(\ell\), which is produced locally and transmitted together with the KV cache.

This approach is vital for edge devices with limited capacity; by identifying when and how much KV cache expands, the system can efficiently control or offload computations without sacrificing fast, incremental generation. Finally, we can define the intermediate output as follows:
\begin{align}
\label{if_total}
    B_{\text{io}}(w, \ell,I_{kv}; \mathcal{Q}^a)
    &= I_{kv}\, B_{kv}(w, \ell, \mathcal{Q}^a)
       + (1-I_{kv})\,T_w\, Q_{a,\ell}.
\end{align}
where \(I_{kv}\) is a factor that determines whether to transmit the KV cache (\(I_{kv} = 1\)) or only the hidden state \((I_{kv} = 0)\). 
In noisy communication scenarios that demand strict latency bounds, sending large KV caches may be infeasible, prompting \(I_{kv}=0\) and thus transmitting only the layer outputs. 
Notably, the KV cache provides significant benefits for speed and server-side efficiency, as it avoids reprocessing all previously generated tokens during autoregressive decoding. However, the KV cache is much larger than transmitting hidden states because it accumulates key and value tensors across multiple layers. Therefore, depending on the communication capacity and latency requirements, sending only the hidden states (smaller but losing the benefits of the cache) may be more advantageous than the entire KV cache. This tradeoff is captured by the binary switch \(I_{kv}\).

Even transmitting layer outputs alone can incur substantial overhead in poor communication conditions. Therefore, in the following subsection, we propose a method to compress these intermediate outputs and decide \(I_{kv}\) accordingly, mitigating such challenges.

\subsection{Adaptive intermediate output compression technique}

\begin{figure}[htbp]
  \centering
  \includegraphics[width=0.5\textwidth]{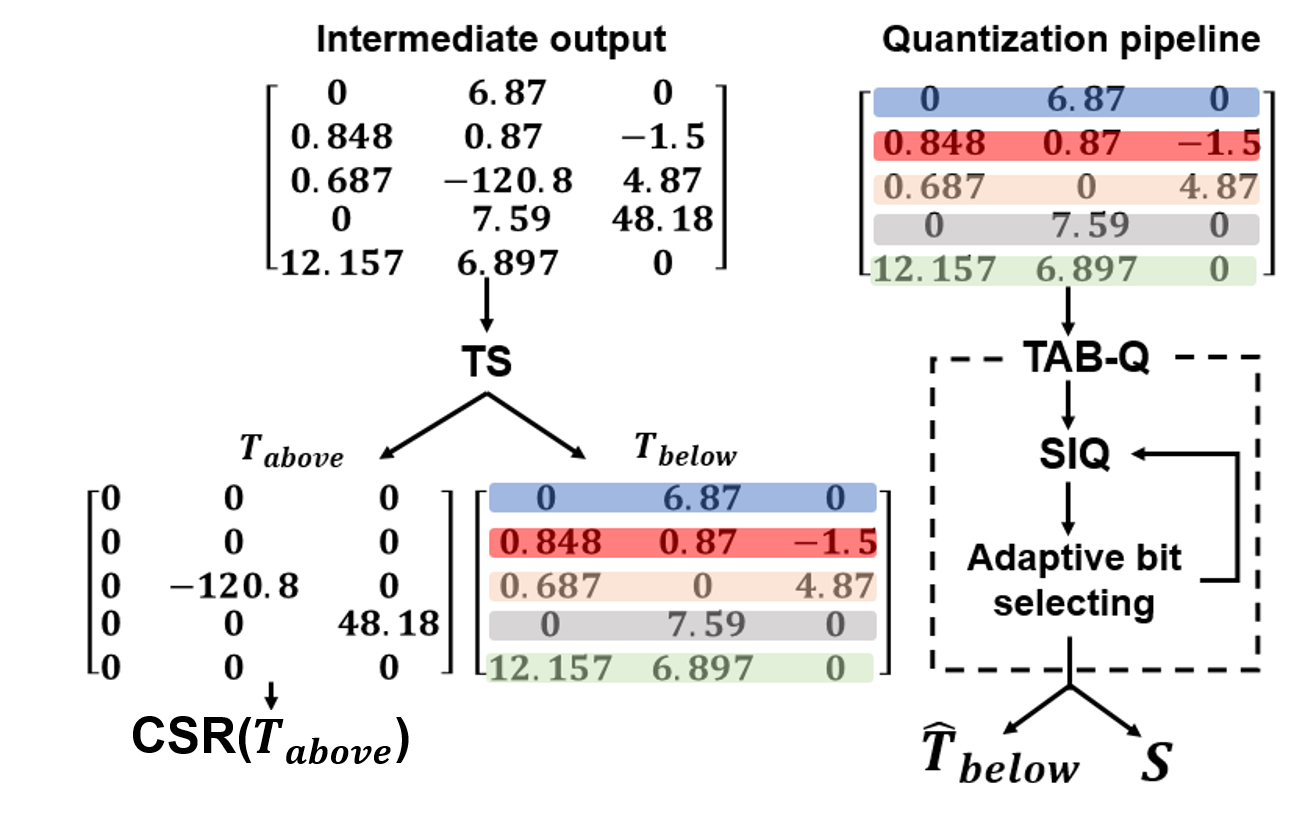}
  \caption{An example of the overall pipeline for applying the proposed intermediate output compression technique.}
  \label{compression}
\end{figure}
Although OPSC already compresses model parameters on the edge device, the intermediate activations generated at the split layer must be further compressed before they are transmitted to the cloud; without this additional reduction, their sheer size would negate the bandwidth and latency advantages of SC. Fig. \ref{compression} presents an example of applying the proposed intermediate output compression technique. The proposed method compresses the intermediate output via a two-stage pipeline: TS and TAB-Q. The two-stage pipeline achieves substantial size reduction while incurring negligible accuracy loss.

\subsubsection{Threshold Splitting}
\begin{figure}[t]
\begin{center}
\centerline{\includegraphics[width=1.0\linewidth]{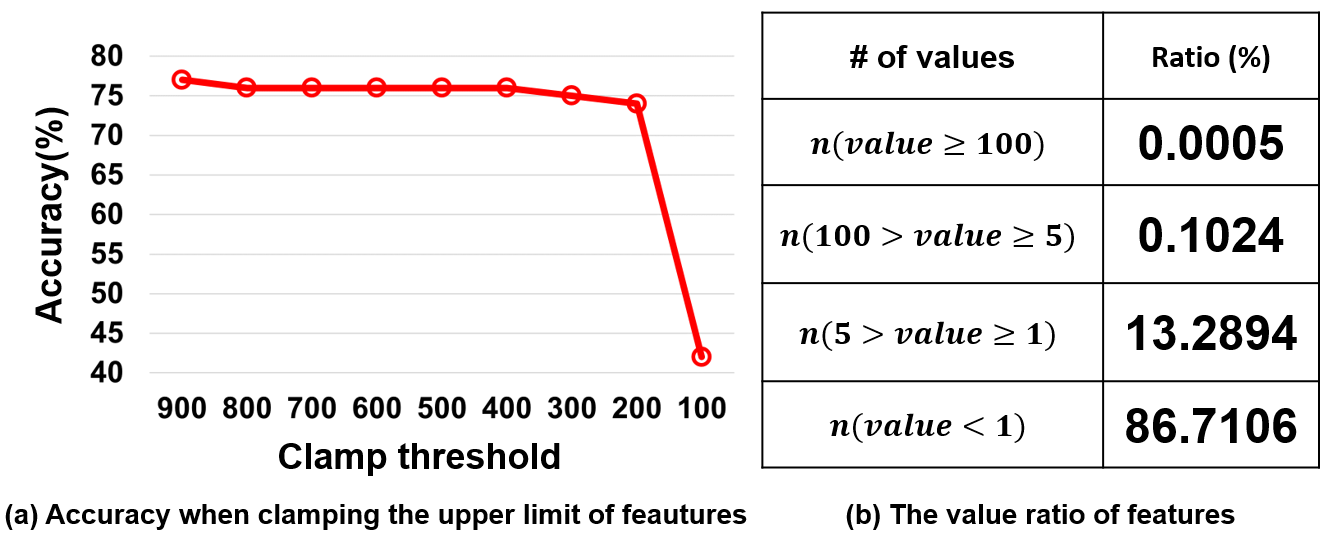}}
\end{center}
\caption{Effect of intermediate output magnitudes-based (Clamping) on the Llama-2 13B model on performance in HellaSwag. (a) Accuracy depending on the upper limit setting of the intermediate output's large value. (b) Distribution of values in intermediate output.}
\label{clamp}
\end{figure}

Because MHA is highly sensitive to outlier activations, quantizing high-magnitude elements can severely degrade accuracy. Fig.~\ref{clamp}(a) illustrates the effect on accuracy when an upper limit is clamped on the intermediate output values to demonstrate the significance of these large values.
A substantial change in accuracy was observed when only the values with absolute magnitudes exceeding 100 in the intermediate output were arbitrarily altered in these examples. This observation suggests that large values in intermediate output drastically influence LLM performance. 
Furthermore, Fig.~\ref{clamp}(b) illustrates the distribution of intermediate output magnitudes. The data reveal that approximately 0.0005\% of the intermediate output values exceed 100, while over 99\% are 100 or below. This distribution suggests that the performance of LLM is highly dependent on a small fraction of values greater than 100, representing only 0.0005\% of the total. As a result, Fig.~\ref{clamp} demonstrates that modifying or restricting these high-magnitude values adversely affects the model's performance.

To preserve large-magnitude values that significantly impact the model, we first introduce the TS pipeline. Initially, the intermediate output\footnote{Whether the KV cache is included or not is determined by $I_{kv}$. However, in the proposed method, the KV cache and layer output are processed separately but in parallel.}, $\mathbf{T}$, is partitioned into \(\mathbf{T}_\text{above}\) and \(\mathbf{T}_\text{below}\) using TS based on a specified threshold, \(\tau\). The formula for TS can be developed as follows:
\begin{align}
\begin{aligned}
\mathbf{T}_{\text{above}} = \mathbf{T}_{ij} \cdot \mathbf{M}_{ij}, \quad \mathbf{T}_{\text{below}} = \mathbf{T}_{ij} \cdot (1 - \mathbf{M}_{ij}), \\ 
\quad \mathbf{M}_{ij} = \begin{cases} 1 & \text{if } |\mathbf{T}_{ij}| \geq \tau \\
0 & \text{otherwise}
\end{cases}
\label{ts}
\end{aligned}
\end{align}
Hence, \(\mathbf{T}_\text{above}\) becomes a highly sparse tensor and is compressed using the compressed sparse row (CSR) format \cite{koza2014compressed}, a widely recognized technique for sparse matrix compression.
The characteristic of CSR indicates that the compression rate is positively correlated to the sparsity of the data (i.e., the higher the sparsity of the data, the higher the compression rate). Therefore, \(\mathbf{T}_\text{above}\) can be transmitted with very low throughput without distortion \cite{du2023predicting}. This separation using TS helps to focus on the most significant values while minimizing the distortion of \(\mathbf{T}_\text{below}\).

\subsubsection{Token-Wise Adaptive Bit Integer Quantization}
The remaining \( \mathbf{T}_{\text{below}} \) is processed through the TAB-Q pipeline. This pipeline implements token-wise operations to preserve the model's ability to differentiate contextual importance. In MHA, different weights are assigned to each token, enabling the model to infer the contextual significance of individual words. By applying operations on a token-wise basis, the relative importance disparities among tokens are maintained throughout the quantization process.

This work introduces the integer quantization (IQ) technique to compress \(\mathbf{T}_{\text{below}}\) efficiently. The IQ technique is widely used for lightweight model deployment because it is straightforward to implement and compatible with most hardware architectures \cite{jacob2018quantization,jiang2024mixtral,frantar2022gptq}.
Specifically, we adopt a asymmetric integer quantization (AIQ) approach, which can be written as follows:
\begin{align}
\mathbf{\hat{T}},\, s,\, z &= \text{AIQ}(\mathbf{T}, Q) = \left\lceil \frac{\mathbf{T}}{s} + z \right\rfloor, \\
s &= \frac{T_{\text{max}}-T_{\text{min}}}{Q_{\text{max}}},\quad
z = \left\lceil \frac{T_{\text{min}}}{s} \right\rfloor,\quad
Q_{\text{max}} = 2^{(Q-1)} - 1.
\label{Proposed}
\end{align}

where \(T_{\text{max}}\) and \(T_{\text{min}}\) are the maximum,minimum value in \(\mathbf{T}\) and \(Q\) represents the quantization level, respectively.

Although IQ is widely applicable, high-variance data can intensify quantization distortion.
In NLP tasks, the distribution of intermediate outputs can vary significantly according to token attention, which complicates selecting a fixed quantization level. This work proposes TAB-Q, an adaptive algorithm that adjusts the quantization level based on the data distribution to address this problem. 
The AIQ approach provides fast computation that is particularly valuable in SC environments while preventing excessive bit usage or severe distortion.

\begin{algorithm}
\caption{TAB-Q}
\begin{algorithmic}[1]
\Require $\mathbf{T}$,  $\bar{Q}$, $\Delta$
\State $\mathbf{T}_{\text{sig}} \gets sign(\mathbf{T})$
\State $\mathbf{\bar{T}} \gets abs(\mathbf{T})$
\State $n \gets W \times HD$ \ \quad /* The number of elements, $\mathbf{T}$

\State $\bar{Q} \gets \bar{Q}-1$
\State $\mathbf{\hat{T}_0}, \mathbf{S_0}, \mathbf{Z_0} \gets \text{AIQ}(\mathbf{\bar{T}},\bar{Q})$

\Repeat
    \State $Q \gets Q - 1$
    \State $\mathbf{\hat{T}}, s, z \gets \text{AIQ}(\mathbf{\bar{T}}, Q)$
    \State $\delta \gets \frac{\sum \left| \left\lfloor \mathbf{\hat{T}_0} / 2^{(\bar{Q} - Q)} \right\rfloor - \mathbf{\hat{T}} \right|}{\text{n}}$
    \If{$\delta > \Delta$}
        \State $\mathbf{\hat{T}^{\ast}} \gets \mathbf{\hat{T}} \odot \mathbf{T}_{\text{sig}}$, $\mathbf{\hat{S}^{\ast}} \gets s$,$\mathbf{\hat{Z}^{\ast}} \gets z$
        \State \textbf{break}
    \EndIf
\Until{$Q < \underline{Q}$}
\State \Return $\mathbf{\hat{T}^{\ast}}, \mathbf{\hat{S}^{\ast}},\mathbf{\hat{Z}^{\ast}}, Q^{\ast}$
\end{algorithmic}
\label{algo}
\end{algorithm}

\noindent
\textbf{Algorithm~\ref{algo}} presents the \emph{TAB-Q} procedure, which adaptively adjusts the quantization level based on the data distribution and a predefined distortion tolerance, \(\Delta\). Below is a line-by-line overview:
\begin{itemize}
    \item \textbf{Lines 1--2 (Sign and Magnitude Extraction):}  
    The input tensor \(\mathbf{T}\) is decomposed into its sign component, \(\mathbf{T}_{\text{sig}}\), and absolute value, \(\mathbf{\bar{T}}\). Handling the sign and magnitude separately helps mitigate quantization distortion for high-variance data.
    
    \item \textbf{Lines 3--4 (Initial Quantization):}  
   The variable \(Q\) is set to \(\bar{Q}-1\) because one bit is reserved for the sign. Then \(\mathbf{\bar{T}}\) is quantized using the maximum level \(\bar{Q}\) to obtain the initial quantized tensor \(\mathbf{\hat{T}_0}\) and scaling factor \(\mathbf{S_0}\).
    
    \item \textbf{Lines 5--9 (Adaptive Bit Reduction and Distortion Check):}  
    The algorithm iteratively decreases \(Q\), reapplies quantization, and measures the distortion \(\delta\). If \(\delta\) \emph{exceeds} \(\Delta\), the loop halts. 
\end{itemize}

\noindent
By terminating as soon as \(\delta\) surpasses \(\Delta\), \textbf{TAB-Q} avoids excessive distortion that could compromise performance. This approach ensures the method preserves both computational and communication efficiency in SC environments. To leverage our complex quantization strategy, we introduce symmetric numeral systems (rANS) \cite{duda2013asymmetric} for encoding, which can efficiently encode multiple quantum variables. Specifically, DietGPU \cite{DietGPU} operates very efficiently because it utilizes GPUs for computation.

In contrast, the compressed intermediate output can be restored on the cloud server as follows:
\begin{align}
\mathbf{\Tilde{T}} = (\mathbf{\hat{T}}^{\ast}_{\text{below}}-\hat{\mathbf{Z}}^{\ast}) \odot \hat{\mathbf{S}}^{\ast} + \mathbf{T}_{\text{above}}.
\label{dequnt}
\end{align}
Recovering the intermediate output is straightforward and efficient, even in the assumed many-to-one scenario, which helps to reduce the load on the cloud server.

\subsection{Selection of Split Layer}
\subsubsection{Maximizing Activation Precision Under Memory Constraints}
\label{{sec:maxQa}}
We now aim to find the split layer \(\ell_w\), the weight-quantization settings \(\mathcal{Q}^w\), and the ``largest'' activation-quantization settings \(\mathcal{Q}^a\) that satisfy both an accuracy constraint and a memory limit. 
Define
\[
\Psi\bigl(\mathcal{Q}^a\bigr) 
\;=\;
\sum_{k=1}^{L} Q_{a,k},
\]
which measures the total activation-bit precision over all layers. Our objective is then to maximize \(\Psi(\mathcal{Q}^a)\). Formally,

\begin{subequations}
\label{eq:MaxQa}
\begin{align}
\bigl(\ell_w^\ast,\;\mathcal{Q}^{w\ast},\;\bar{\mathcal{Q}^{a}}\bigr)
\;=\;
& \arg\max_{\ell_w,\,\mathcal{Q}^w,\,\mathcal{Q}^a}
  \;\;
  \Psi\bigl(\mathcal{Q}^a\bigr), \\[4pt]
\text{s.t:}&
\quad 
A\bigl(\ell_w, \mathcal{Q}^w, \mathcal{Q}^a\bigr)
\;\ge\;
A_{\text{base}} - A_{\Delta},
\label{eq:maxQaAccConstraint}\\[3pt]
&
\quad
M\bigl(\ell_w, \mathcal{Q}^w\bigr)
\;+\;
B_{kv}\bigl(\bar{W}, \ell;\, \mathcal{Q}^a\bigr)
\;\le\;
\mathcal{M}.
\label{eq:maxQaMemConstraint}
\end{align}
\end{subequations}

\noindent
Here, \(A_{\text{base}} - A_{\Delta}\) denotes the lower bound for acceptable accuracy, ensuring performance does not drop by more than \(A_{\Delta}\). The constraint  \eqref{eq:maxQaMemConstraint} ensures that the total memory usage  \(M\bigl(\ell_w, \mathcal{Q}^w\bigr) + B_{kv}\bigl(\bar{W}, \ell;\,\mathcal{Q}^a\bigr)\) does not exceed \(\mathcal{M}\). Note that \(\bar{W}\) is treated as \emph{fixed}: it corresponds to the maximum number of tokens the edge device is expected to generate and thus must be fully accommodated under the memory limit. The solution \(\bigl(\ell_w^\ast,\;\mathcal{Q}^{w\ast},\;\bar{\mathcal{Q}^a}\bigr)\) provides the configuration that maximizes overall activation precision without violating the accuracy or memory constraints.

\paragraph{Solution Approach} 
Since \(\ell_w\), \(\mathcal{Q}^w\), and \(\mathcal{Q}^a\) typically come from 
discrete sets (e.g., bitwidths 4, 8, 16), one can:
\begin{enumerate}
    \item Set $w$ to \(\bar{W}\) given maximum feasible token length (i.e., the largest token length we want to support on the edge device).
    \item \emph{Enumerate} all possible \(\ell_w\), \(\mathcal{Q}^w\), and \(\mathcal{Q}^a\).
    \item \emph{Check} each candidate configuration against constraints 
          \eqref{eq:maxQaAccConstraint}--\eqref{eq:maxQaMemConstraint}.
    \item \emph{Select} the combination 
          \(\bigl(\ell_w^\ast,\;\mathcal{Q}^{w\ast},\;\bar{\mathcal{Q}^a}\bigr)\)
          that yields the largest \(\Psi(\mathcal{Q}^a)\).
\end{enumerate}

\noindent
Because \(\bar{T}_{w}\) is fixed and thus not minimized, the solution ensures the edge device can handle the \emph{full} token length \(\bar{T}_w\) without running out of  memory, while still maximizing the activation precision and maintaining accuracy  within \(\Delta\).

\subsubsection{Early Exit Strategy for Delay Constraints}
\paragraph{Objective Function Setting}
While Eq~\eqref{eq:MaxQa} focuses on satisfying memory constraints without compromising accuracy, some applications further demand strict end-to-end latency guarantees. Let \(D\) denote the maximum allowable delay for completing a single inference. To analyze communication overhead, we adopt an \(\varepsilon\)-outage reliability framework, in which the worst-case latency for transmitting data of size \(D_{\rm tx}\) at rate \(R\) is given~by:
\begin{align}
\mathcal{L}_{\varepsilon}(D_{\rm tx}; R)
&= \frac{D_{\rm tx}}{R}\,
\left\lceil \frac{\ln(\varepsilon)}{\ln\Bigl(P_o(R)\Bigr)} \right\rceil, \label{comm_latency}\\[1mm]
\text{where}\quad P_o(R)
&= 1 - \exp\!\Biggl(-\frac{2^{\frac{R}{W}} - 1}{\gamma}\Biggr).
\end{align}

Here, \(\varepsilon>0\) is the target outage probability, \(W\) is the available bandwidth, and \(\gamma\) is the received signal-to-noise ratio (SNR).

On the computation side, let \(\mathcal{L}_{c}(w)\) denote the local (on-device) processing time required for \(w\) tokens\footnote{%
Local computation latency was profiled in real time on the target edge device.}. Combining both terms, the total edge-device inference latency when generating the \(w\)-th token up to layer \(\ell\) is:
\begin{align}
\label{total}
\mathcal{L}_{t}\Bigl(T_{w}, \ell, \mathcal{Q}^{a}, I_{kv}; R\Bigr)
&= \underbrace{\mathcal{L}_{c}(w)}_{\substack{\text{Local}\\\text{computation}}} \nonumber\\[1mm]
&\quad + \underbrace{\mathcal{L}_{\epsilon}\!\Bigl(B_{\text{io}}(w, \ell, I_{kv}; \mathcal{Q}^{a}),\, R\Bigr)}_{\substack{\text{communication}}}.
\end{align}

To ensure timely inference, we impose the constraint
\(
\mathcal{L}_{t}\bigl(T_{w}, \ell, \mathcal{Q}^{a}, I_{kv}; R\bigr) 
\,\le\, 
D.
\)
Note that, because the cloud server’s computation and queueing latencies fluctuate with the instantaneous number of active clients—making precise analytical modeling impractical—the server instead communicates to each edge device a load-aware deadline that implicitly reflects its current operating state.
We now formalize an \emph{early-exit} objective, which permits the system to reduce the number of generated tokens or skip specific layers whenever the latency constraint would otherwise be violated. Let \(\bar{W}\) be the maximum number of tokens we are willing to generate. We define the following objective: 
\begin{equation}
\label{eq:EE_obj}
\max_{\substack{0 \,\le\, w \,\le\, \bar{W}\\1 \,\le\, \ell \,\le\, L}} 
\;\;
w\,\ell
\quad
\text{subject to}
\quad
\mathcal{L}_{t}\bigl(T_{w}, \ell, \mathcal{Q}^{a}, I_{kv};R\bigr)
\,\le\,
D.
\end{equation}
Here, \(w \times \ell\) measures the total inference “depth” (in terms of tokens and layers) that can be processed before exceeding the target latency~\(D\). Larger values of \(w \times \ell\) typically correspond to higher-quality outputs.

\paragraph{Optimization}
Notice that \(\mathcal{L}_{\epsilon}(\cdot,R)\) in~\eqref{comm_latency} is non-monotonic in~\(R\). Following the approach in~\cite{yun2022cooperative}, define the function
\[
g(R) 
\;=\; 
\tfrac{\ln\bigl(1/P_{o}(R)\bigr)}{R},
\]
and constrain \(R\) to lie in a feasible interval \([\underline{R},\,\overline{R}]\). Then the optimal rate \(R^*\) that minimizes \(g(\cdot)\) and hence yields the smallest worst-case communication delay is found via:
\begin{equation}
\label{eq:epsilon_rstar}
R^* 
\;=\; 
\arg\min_{\,R \in [\underline{R}, \,\overline{R}]}
\;g(R).
\end{equation}
In practice, the solution \(R^*\) can be found by simple one-dimensional search.

Having solved~\eqref{eq:epsilon_rstar} for \(R^*\), we then use~\eqref{eq:EE_obj} to determine how many tokens \(w\) and which layers \(\ell\) can be processed while keeping total latency below~\(D\). As soon as \(\mathcal{L}_{t}(\cdot)\) threatens to exceed~\(D\), an early-exit decision is triggered: the system either reduces the number of tokens, disables KV caching (\(I_{kv}=0\)), or compresses the intermediate outputs more aggressively.

\paragraph{Algorithmic Solution}
Algorithm~\ref{alg:earlyexit} details our proposed approach, which integrates the memory-feasible configuration 
\(\bigl(\ell_w^{\ast}, \,\mathcal{Q}^{w\ast},\,\bar{\mathcal{Q}}^{a}\bigr)\) 
obtained from \eqref{eq:MaxQa} with a real-time monitoring of the total latency~\(\mathcal{L}_{t}\). As the model processes each token, we evaluate whether sending the intermediate outputs (possibly including the KV cache) at the current precision will exceed the time limit~\(D\). If it does, we apply additional compression steps (e.g., skipping KV caching or reducing token count) until the latency is once again under control. By adaptively trading off token generation depth and intermediate-output compression, the algorithm ensures adherence to strict delay budgets even under time-varying network~conditions.

\begin{algorithm}[t]
    \caption{Early Exit Strategy under Delay Constraints}
    \label{alg:earlyexit}
    \begin{algorithmic}[1]
    \Require \(\bigl(\ell_w^\ast,\;\mathcal{Q}^{w\ast},\;\bar{\mathcal{Q}}^{a}\bigr)\) 
      satisfying~\eqref{eq:MaxQa} for a maximum token count \(\bar{W}\).\\
      Delay tolerance \(D\).\\
      Latency function \(\mathcal{L}_{t}\bigl(T_{w},\ell,\bar{\mathcal{Q}}^{a},I_{kv};\,R\bigr)\) 
      from~\eqref{total}.\\
    \Ensure 
      \(w^\ast \le \bar{W}\) maximizing \(w \times \ell\) s.t \(\mathcal{L}_{t} \le D\).
    \Statex
    \State \textbf{Compute} \(R^{\ast}\) using~\eqref{eq:epsilon_rstar}
    \State \(w \gets 0,\; I_{kv} \gets 1\)
    \For{\(w = 1\) to \(\bar{W}\)}
        \For{\(\ell = 1\) to \(L\)}
            \State \text{Forward pass up to layer \(\ell\) for token~\(w\)}
            \State \(\text{latency} \gets \mathcal{L}_{t}\bigl(T_{w},\ell,\bar{\mathcal{Q}}^{a},I_{kv};\,R^{\ast}\bigr)\)
            \If{\(\text{latency} > D\)}
                \State \(\hat{T}_{w} \gets \textsc{TabQ}(T_{w};\,I_{kv})\) 
                       \quad\Comment{Compress}
                \State \(\text{latency} \gets \mathcal{L}_{t}\bigl(\hat{T}_{w},\ell,\bar{\mathcal{Q}}^{a},I_{kv};\,R^{\ast}\bigr)\) 
                \If{\(\text{latency} \le D\)}
                    \State \textbf{return} \(\hat{T}_{w}\) \quad\Comment{Early exit successful}
                \Else
                    \State \(I_{kv} \gets 0\)
                    \State \(\hat{T}_{w} \gets \textsc{TabQ}(T_{w};\,I_{kv})\)
                    \State \(\text{latency} \gets \mathcal{L}_{t}\bigl(\hat{T}_{w},\ell,\bar{\mathcal{Q}}^{a},I_{kv};\,R^{\ast}\bigr)\)
                    \If{\(\text{latency} > D\)}
                        \While{\(\text{latency} > D\)}
                            \State \(w \gets w - 1\) \quad\Comment{Reduce token}
                            \State \(\text{latency} \gets \mathcal{L}_{t}\bigl(\hat{T}_{w},\ell,\bar{\mathcal{Q}}^{a},I_{kv};\,R^{\ast}\bigr)\)
                        \EndWhile
                        \State \textbf{return} \(\hat{T}_{w}\) \quad\Comment{Early exit}
                    \EndIf
                \EndIf
            \EndIf
        \EndFor
    \EndFor
    \State \(\hat{T}_{w} \gets \textsc{TabQ}(T_{w};\,I_{kv})\)
    \State \textbf{return} \(\hat{T}_{w}\)
    \end{algorithmic}
\end{algorithm}

\section{Evaluation}

\subsection{Experimental setup}
We validate the proposed SC method on two Llama2 variants---7B-hf (7B) and 13B-hf (13B) \cite{touvron2023llama}---each comprising 32 and 40 decoder layers, respectively. In this work, we treat all decoder layers as a single “splittable” stack, allowing the split point \(\ell\) to vary from 1 to 32 (for 7B) or 1 to 40 (for 13B). Unless otherwise noted, we fix the split layer at \(\ell=20\), the threshold at \(\tau=5\), the acceptable distortion \(A_{\Delta}=1\%\), and \(\Delta=0.2\), which balances accuracy and communication overhead based on our preliminary tests. Note that the clip for activation in OPSC settings is not used.
For the communication experiments, we set \(\varepsilon=0.001\), \(W=10\)~MHz, \(\sigma_h^2=1\), and \(\gamma=10\). We run the edge inference on a Jetson Xavier NX (16\,GB) and the cloud inference on an A6000 GPU.
All evaluations are conducted in a zero-shot setting on HellaSwag (HS) \cite{zellers2019hellaswag}, PIQA \cite{bisk2020piqa}, ARC-e/c \cite{clark2018arc}, BoolQ \cite{clark2019boolq}, and Winogrande (Wino.) \cite{sakaguchi2021winogrande}. Models are assessed using pretrained weights and fixed prompt templates without task-specific fine-tuning or in-context demonstrations. This experimental design isolates the effects of the proposed split-compression framework from confounding factors, reflects realistic deployment scenarios where edge devices cannot fine-tune large models, and ensures fair comparisons across benchmark tasks.

\subsection{Performance comparison}

\begin{figure}[htbp]
  \centering
  \includegraphics[width=0.5\textwidth]{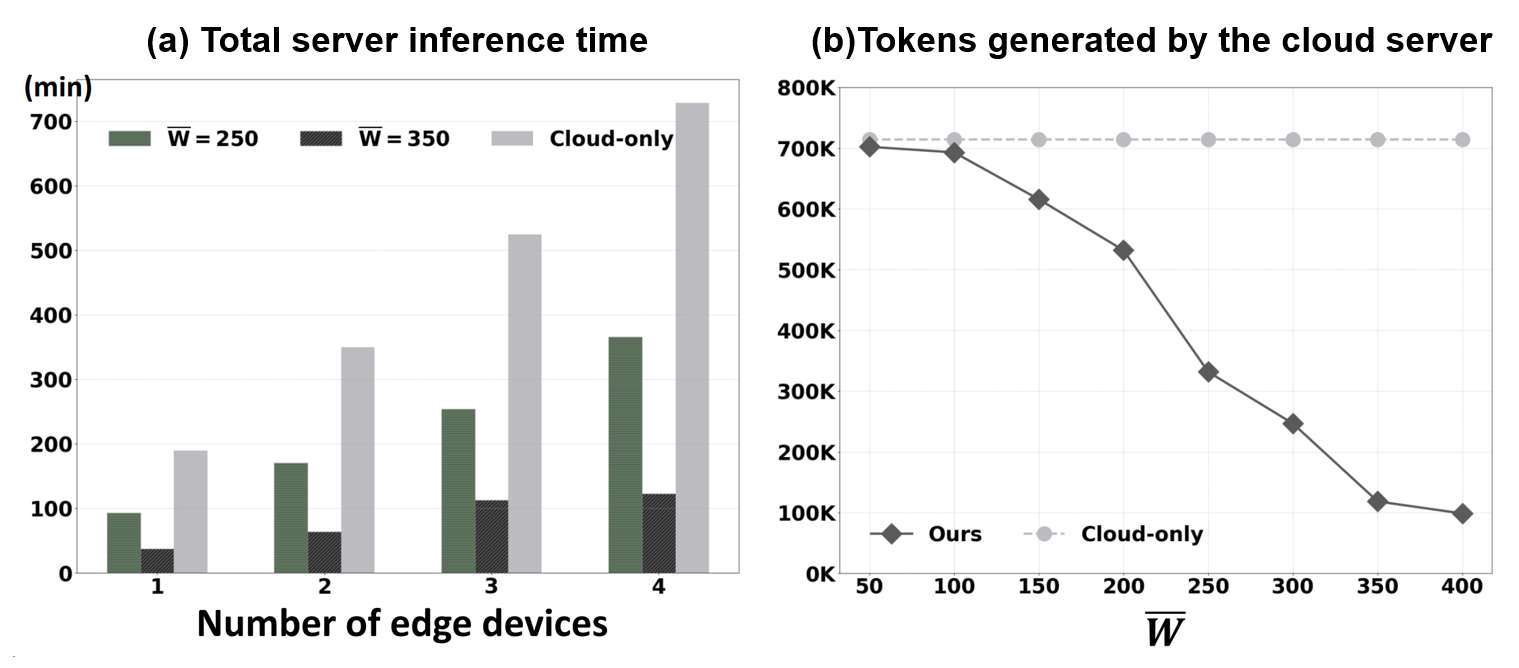}
  \caption{\textbf{(a)}~Total server inference time (in minutes) versus the number of edge devices for three configurations: 
    `Cloud-only' (all tokens processed by the server) and our SC method with $\bar{W} = 250$ and $\bar{W} = 350$. 
    \textbf{(b)}~Number of tokens generated by the server as $\bar{W}$ varies. 
    Our approach gradually offloads more inference steps to the edge device, significantly reducing both server inference time and token generation overhead.}
  \label{fig:server_time_tokens}
\end{figure}

Fig.~\ref{fig:server_time_tokens}(a) displays the server's total inference time as the number of edge devices increases. We compare a baseline in which all tokens are processed by the server ("Cloud-only") against two SC variants with maximum sequence lengths of $\bar{W}=250$ or $\bar{W}=350$ on the edge devices. As the number of connected edge devices increases, the proposed SC approach maintains a lower server workload than the Cloud-only scheme, demonstrating superior scalability. 
We also note in Fig.~\ref{fig:server_time_tokens}(a) that the server inference time exhibits nonlinear growth with increasing edge device count. This nonlinear behavior stems from server-side bottlenecks, including dynamic batching overhead, queueing delays, and GPU memory management constraints, which collectively intensify computational overhead under high concurrency conditions. Despite these practical limitations, the proposed SC framework consistently achieves lower server latency than the cloud-only baseline, demonstrating its scalability benefits in realistic multi-user deployments.
Fig.~\ref{fig:server_time_tokens}(b) further illustrates the number of tokens generated on the server for different values of $\bar{W}$. The proposed method maximizes the on-board inference at the edge, significantly reducing the number of tokens the server must generate. In contrast, the Cloud-only approach saturates at a high token count irrespective of $\bar{W}$. These findings demonstrate that the proposed framework effectively reduces server-side overhead while preserving inference quality.

\begin{figure}[htbp]
  \centering
  \includegraphics[width=0.5\textwidth]{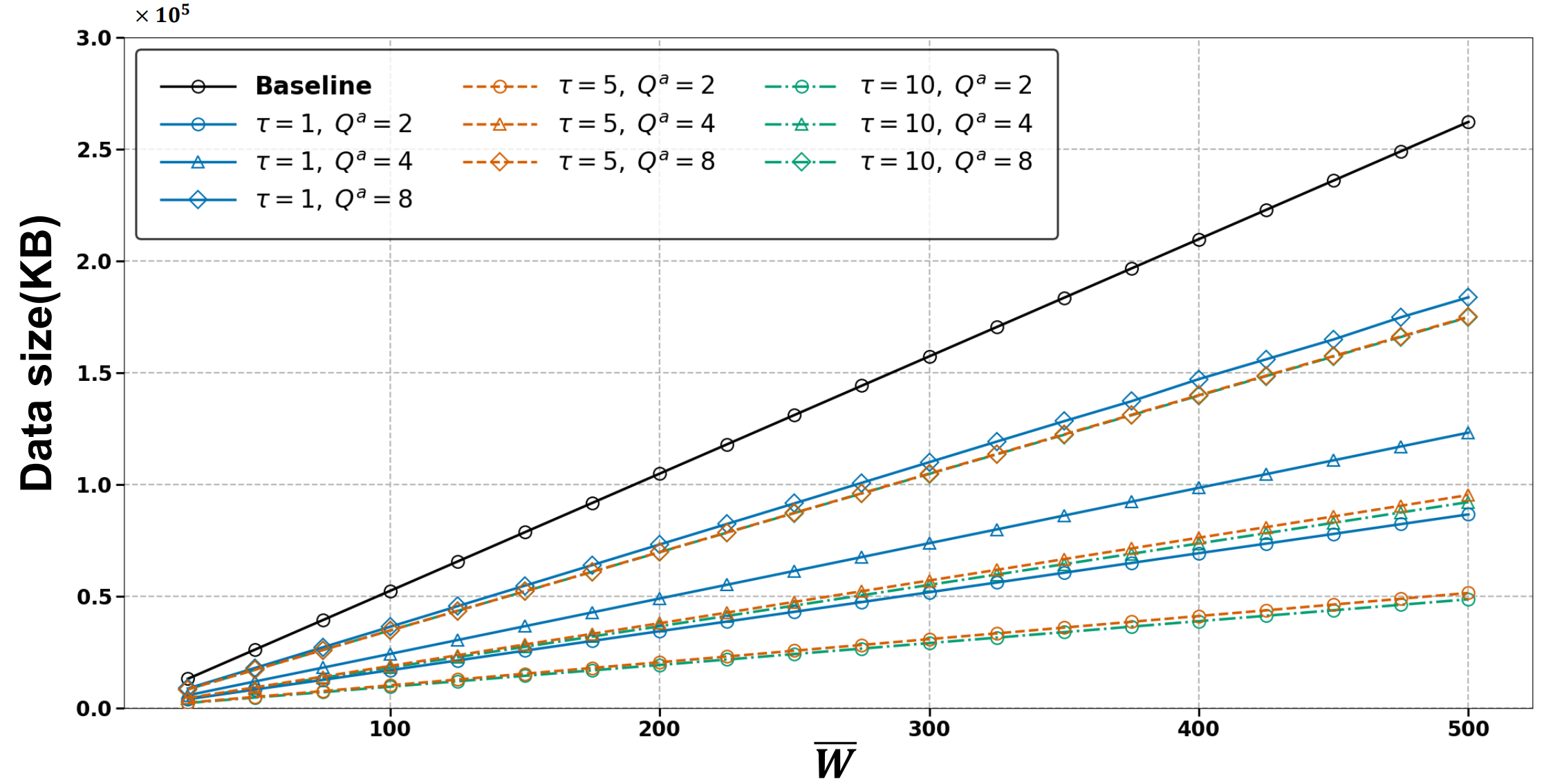}
  \caption{Data size of intermediate outputs versus token length $W$, comparing our proposed compression approach for various threshold values $\tau \in \{1,5,10\}$ and maximum activation bits $\bar{Q}^{a} \in \{2,4,8\}$. The ``Baseline'' line (black circles) denotes no compression, resulting in the largest data size.}
  \label{fig:intermediate_compression}
\end{figure}

Fig.~\ref{fig:intermediate_compression} illustrates the relationship between intermediate output size (in kilobytes) and the maximum token length $\bar{W}$ across various threshold ($\tau$) and maximum activation-bit ($\bar{Q}^a$) settings. The Baseline curve corresponds to transmitting the uncompressed hidden states or KV caches, yielding the most significant data size. In contrast, the two-stage compression framework (TS + TAB-Q) reduces the data size substantially by isolating the large-magnitude values (via $\tau$) and then performing token-wise integer quantization (via $\bar{Q}^a$). The higher $\tau$ values filter out larger elements, resulting in increased sparsity, whereas lower $\bar{Q}^a$ values enforce more aggressive quantization. Therefore, varying $\tau$ and $\bar{Q}^a$ allows flexible control over communication overhead, making it feasible to process extended token sequences with minimal data transfer requirements.

Table~\ref{tab:memory_lim} compares zero-shot performance when deploying the 7B model under tight memory constraints on the edge device. We contrast two schemes: (i)~fully lightweight quantization using Atom~\cite{zhao2024atom}, and 
(ii)~the proposed SC method with $\bar{W}=50$, employing the adaptive intermediate output compression at the split layer ($\tau=5$, $\bar{Q}^{a}=4$). We tested multiple split layers~($\ell$) in each setting. Overall, the proposed approach achieves higher accuracy than Atom across all datasets and split layers. By compressing only the portion of the model that must reside on the edge device while offloading the remainder to the cloud, the proposed framework mitigates the accuracy degradation often caused by aggressive quantization. Our method outperforms Atom through three advantages: TS preserves critical values without quantization loss versus Atom's 8-bit outlier quantization; TAB-Q enables per-token adaptation versus static quantization; and SC architecture allows cloud-side full-precision computation.

\begin{table}[htb!]
    \centering
    \caption{Performance comparison across different split layers.}
    \label{tab:memory_lim}
    \begin{tabular}{ccccccc}
        \toprule
        $\ell$ & Method & PIQA & ARC-e & BoolQ & HS & Wino. \\
        \toprule
        \multirow{2}{*}{5} & Atom      & 75.63 & 52.69 & 67.71 & 67.18 & 64.48 \\
                           & \cellcolor{lightgray}Ours & \cellcolor{lightgray}76.50 & \cellcolor{lightgray}54.12 & \cellcolor{lightgray}70.67 & \cellcolor{lightgray}70.73 & \cellcolor{lightgray}67.32 \\
        \midrule
        \multirow{2}{*}{10} & Atom      & 74.86 & 51.73 & 65.84 & 66.57 & 63.69 \\
                            & \cellcolor{lightgray}Ours & \cellcolor{lightgray}76.01 & \cellcolor{lightgray}52.15 & \cellcolor{lightgray}68.84 & \cellcolor{lightgray}68.35 & \cellcolor{lightgray}66.38 \\
        \midrule
        \multirow{2}{*}{15} & Atom      & 75.08 & 51.43 & 67.98 & 67.81 & 64.56\\
                            & \cellcolor{lightgray}Ours & \cellcolor{lightgray}76.33 & \cellcolor{lightgray}52.44 & \cellcolor{lightgray}69.60 & \cellcolor{lightgray}68.57 & \cellcolor{lightgray}65.27 \\
        \midrule
        \multirow{2}{*}{20} & Atom      & 75.73 & 52.40 & 67.25 & 68.23 & 65.04 \\
                            & \cellcolor{lightgray}Ours & \cellcolor{lightgray}76.17 & \cellcolor{lightgray}53.37 & \cellcolor{lightgray}67.22 & \cellcolor{lightgray}68.63 & \cellcolor{lightgray}65.11 \\
        \midrule
        \multirow{2}{*}{25} & Atom      & 75.63 & 52.69 & 67.28 & 68.56 & 65.04 \\
                            & \cellcolor{lightgray}Ours & \cellcolor{lightgray}75.79 & \cellcolor{lightgray}53.11 & \cellcolor{lightgray}68.17 & \cellcolor{lightgray}68.65 & \cellcolor{lightgray}65.90 \\
        \midrule
        \multirow{2}{*}{30} & Atom      & 76.06 & 52.40 & 67.19 & 68.49 & 64.40 \\
                            & \cellcolor{lightgray}Ours & \cellcolor{lightgray}76.12 & \cellcolor{lightgray}52.57 & \cellcolor{lightgray}67.31 & \cellcolor{lightgray}68.55 & \cellcolor{lightgray}64.25 \\
        \bottomrule
    \end{tabular}
\end{table}

Table~\ref{tab:quantized_llama} shows the accuracy of our proposed SC compression strategy versus three established quantization techniques---SmoothQuant (E1)~\cite{xiao2023smoothquant}, OmniQuant (E2)~\cite{shao2023omniquant}, and Atom (E3)~\cite{zhao2024atom}---for both 7B and 13B models. All approaches apply $Q_{w1}=4$ and $Q_{w2}=4$ for weight quantization; the table reports results under different activation-bit settings, $\bar{Q}^{a}\in\{3,4\}$. Unlike traditional activation quantization methods developed for general inference settings, the proposed approach  more effectively applies quantization at the split layer, making it highly suitable for SC environments. The crucial advantage of the proposed method is its ability to balance the tradeoff between minimizing communication delay and maximizing accuracy in edge–cloud systems. Strict delay requirements are met without compromising accuracy by applying intermediate output compression, even in resource-constrained environments.

\begin{table}[htb!]
    \centering
    \caption{Comparison with different LLM compression techniques.}
    \label{tab:quantized_llama}
    \setlength{\tabcolsep}{4.5pt}
    \begin{tabular}{cccccccc}
        \toprule
         $\bar{Q}^{a}$ & Method & PIQA & ARC-e & ARC-c & BoolQ & HS & Wino. \\
        \toprule
        \multicolumn{8}{c}{7B} \\ 
        \midrule
        \multirow{4}{*}{3} & E-1 & 51.88 & 29.33 & 31.84 & 46.07 & 29.01 & 50.63 \\
        & E-2 & 53.05 & 30.70 & 30.77 & 38.53 & 28.59 & 51.03 \\
        & E-3 & 67.79 & 43.73 & 31.40 & 62.35 & 55.26 & 57.38 \\
        & \cellcolor{lightgray}Ours & \cellcolor{lightgray}75.19 & \cellcolor{lightgray}52.65 & \cellcolor{lightgray}38.05 & \cellcolor{lightgray}73.21 & \cellcolor{lightgray}69.22 & \cellcolor{lightgray}63.30 \\
        \midrule
        \multirow{4}{*}{4} & E1 & 62.30 & 40.00 & 31.37 & 59.32 & 43.16 & 47.00 \\
                           & E2 & 65.30 & 45.17 & 30.94 & 64.43 & 56.16 & 47.56 \\
                           & E3 & 75.46 & 51.43 & 38.91 & 68.47 & 69.67 & 63.30 \\
        & \cellcolor{lightgray}Ours & \cellcolor{lightgray}76.33 & \cellcolor{lightgray}54.63 & \cellcolor{lightgray}40.44 & \cellcolor{lightgray}74.43 & \cellcolor{lightgray}70.77 & \cellcolor{lightgray}67.09 \\
        \midrule
        \multicolumn{8}{c}{13B} \\ 
        \midrule
        \multirow{4}{*}{3} & E1 & 48.25 & 27.18 & 29.12 & 48.85 & 25.65 & 51.15 \\
                           & E2 & 50.49 & 27.67 & 29.30 & 39.40 & 25.77 & 52.63 \\
                           & E3 & 68.28 & 47.22 & 34.64 & 65.11 & 60.79 & 56.43 \\
                           & \cellcolor{lightgray}Ours & \cellcolor{lightgray}70.46 & \cellcolor{lightgray}49.58 & \cellcolor{lightgray}35.49 & \cellcolor{lightgray}66.09 & \cellcolor{lightgray}62.92 & \cellcolor{lightgray}58.48 \\
        \midrule
        \multirow{4}{*}{4} & E1 & 64.15 & 40.50 & 30.52 & 62.29 & 46.75 & 50.92 \\
                           & E2 & 65.35 & 45.97 & 32.71 & 62.84 & 59.05 & 54.96 \\
                           & E3 & 77.31 & 55.85 & 42.41 & 67.52 & 73.88 & 67.48 \\
                           & \cellcolor{lightgray}Ours & \cellcolor{lightgray}78.07 & \cellcolor{lightgray}57.49 & \cellcolor{lightgray}43.33 & \cellcolor{lightgray}70.15 & \cellcolor{lightgray}74.76 & \cellcolor{lightgray}69.69 \\
        \bottomrule
    \end{tabular}
\end{table}

Further, the Tables~\ref{tab:memory_lim} and \ref{tab:quantized_llama} demonstrate that the proposed framework consistently outperforms Atom under identical memory constraints. This performance advantage results from the split-aware architecture, where only the front-end layers executed on edge devices undergo quantization, while the back-end layers maintain complete precision on the server. This design preserves semantic fidelity in the most critical stages of inference. In contrast, Atom applies uniform quantization across the entire model, including accuracy-sensitive final layers, thereby introducing additional distortion that degrades overall performance.

To evaluate the effectiveness of our OPSC quantization strategy under different \(\mathcal{Q}^w\) configurations, we conducted detailed tests on both the front- and back-end portions of the Llama2 model. Table~\ref{tab:split_quant} summarizes the perplexity\footnote{Perplexity measures how well a language model predicts a given text. Lower perplexity indicates better predictive performance, meaning the model is less ``confused'' about the next word.} on the WikiText2 and C4 datasets for the 7B and 13B variants. Quantifying fewer layers in the front-end method yields lower perplexity, implying that the final layers are susceptible to precision reduction. Quantizing the back-end (i.e., the last layers) generally produces slightly higher perplexity at the same $\ell_w$, confirming that later layers play a more critical role in accurate language modeling. For both 7B and 13B models, increasing $\ell_w$ leads to progressively higher perplexity, as more of the network is placed under 4-bit quantization. These findings highlight the importance of strategically selecting which layers to compress to balance memory savings with minimal performance loss in SC scenarios.

\begin{table}[htb!]
\caption{Perplexity of quantized Llama models with split layer quantization on WikiText2 and C4 datasets}
\label{tab:split_quant}
\centering
\begin{tabular}{c|c|cc||c|cc}
\toprule
 & \textbf{$\ell_w$} & \multicolumn{2}{c||}{front-end method (↓)} & \textbf{$\ell_w$} & \multicolumn{2}{c}{back-end method(↓)} \\
\cmidrule{3-4}\cmidrule{6-7}
& & Wiki & C4 & & Wiki & C4 \\
\midrule
\multirow{9}{*}{7B} & 4 & 5.538 & 7.030 & 4 & 5.607 & 7.123 \\
& 8 & 5.608 & 7.115 & 8 & 5.694 & 7.222 \\
& 12 & 5.682 & 7.206 & 12 & 5.781 & 7.338 \\
& 16 & 5.739 & 7.302 & 16 & 5.831 & 7.427 \\
& 20 & 5.840 & 7.435 & 20 & 6.074 & 7.684 \\
& 24 & 5.920 & 7.562 & 24 & 6.394 & 7.999 \\
& 28 & 5.997 & 7.673 & 28 & 6.566 & 8.160 \\
& 32 & 6.030 & 7.755 & 32 & 6.103 & 7.818 \\
\midrule
\multirow{11}{*}{13B} & 4 & 4.915 & 6.496 & 4 & 4.950 & 6.531 \\
& 8 & 4.956 & 6.537 & 8 & 5.642 & 7.429 \\
& 12 & 5.000 & 6.585 & 12 & 5.032 & 6.623 \\
& 16 & 5.049 & 6.643 & 16 & 5.094 & 6.694 \\
& 20 & 5.096 & 6.705 & 20 & 5.172 & 6.784 \\
& 24 & 5.139 & 6.766 & 24 & 5.357 & 7.019 \\
& 28 & 5.180 & 6.821 & 28 & 5.531 & 7.281 \\
& 32 & 5.222 & 6.874 & 32 & 5.642 & 7.429 \\
& 36 & 5.261 & 6.926 & 36 & 5.704 & 7.468 \\
& 40 & 5.312 & 6.991 & 40 & 5.312 & 6.991 \\
\bottomrule
\end{tabular}
\end{table}

\begin{figure}[htbp]
  \centering
  \includegraphics[width=0.5\textwidth]{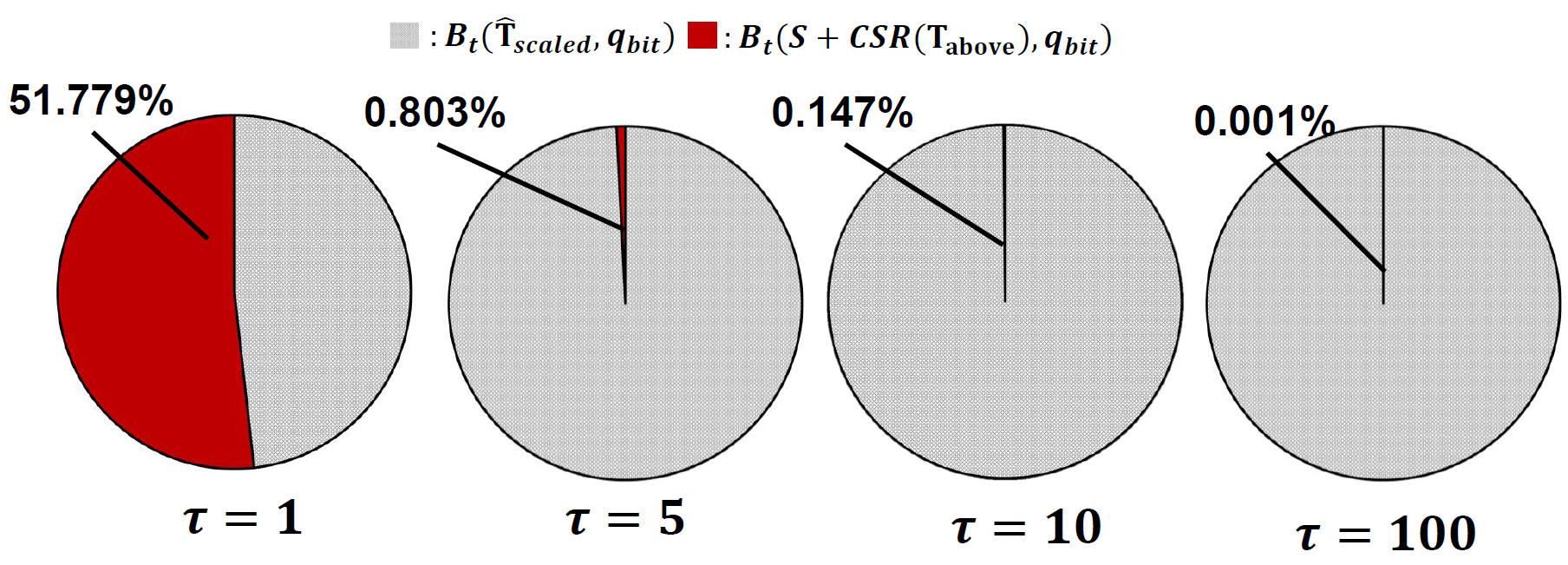}
  \caption{Data size ratio of \( \mathbf{\hat{T}}_{\text{below}} \) (gray) and \( \mathbf{T}_{\text{above}} \) (red) according to $\tau$}
  \label{ratio}
\end{figure}

Fig.~\ref{ratio} illustrates how the required data size of intermediate outputs varies with \(\tau\), and how it contributes to \( \mathbf{T}_{\text{below}} \) and \( \mathbf{T}_{\text{above}} \). When \(\tau\) is 1, the cost of compressing and transmitting \( \mathbf{T}_{\text{above}} \) is high, resulting in a reduced compression ratio. However, when \(\tau\) exceeds 1, the sparsity of \( \mathbf{T}_{\text{above}} \) increases significantly. Consequently, the impact on the transmission latency of \( \mathbf{T}_{\text{above}} \) becomes negligible.



Table~\ref{tab:ablation_performance} evaluates the impact of the proposed two-stage compression by comparing the 13B model with TAB-Q alone against the combined TS + TAB-Q design. The baseline involves no intermediate output compression, whereas `Baseline + TAB-Q' applies quantization alone. Although TAB-Q significantly reduces accuracy when used in isolation (e.g., HellaSwag drops from 77.31\%  to 45.26\%), adding TS (`Baseline + TS + TAB-Q') restores performance nearly to baseline levels. This observation demonstrates that TS effectively preserves large-magnitude values critical to the model, mitigating the distortion introduced by TAB-Q. 

\begin{table}[htb!]
    \centering
    \caption{Ablation study on our proposed method in 13B}
    \begin{adjustbox}{max width=\columnwidth}
    \label{tab:ablation_performance}
    \begin{tabular}{lcccc}
        \toprule
        Ablation study & HS & ARC-e & ARC-c & PIQA \\
        \midrule
        Baseline & 77.31 & 79.12 & 43.96 & 78.89 \\
        Baseline+TAB-Q & 45.26 & 33.68 & 24.50 & 55.60 \\
        Baseline+TS+TAB-Q & 77.09 & 76.61 & 43.62 & 78.02 \\
        \bottomrule
    \end{tabular}
    \end{adjustbox}
\end{table}

Further, to evaluate the cross-model generalization capability of the proposed framework, 
Table~\ref{tab:llm_comparison} compares baseline performance \cite{qwen25,mistral_nemo,meta_llama_31,abdin2024phi} 
with our method across four distinct LLMs on five benchmark tasks (ARC-e, ARC-c, BoolQ, 
HellaSwag, and Wino). Values in \textcolor{red}{red} denote improvements over the baseline, 
while those in \textcolor{blue}{blue} indicate performance drops.

\begin{table}[htb!]
\centering
\caption{Benchmark results for various LLMs across multiple tasks. For each model, the \emph{first row} lists its baseline performance, whereas the \emph{second row} shows results after applying our proposed method.}
\label{tab:llm_comparison}
\begin{adjustbox}{max width=\columnwidth}
\setlength{\tabcolsep}{4pt}
\begin{tabular}{lccccc}
\toprule
\textbf{Model} & \textbf{ARC-e} & \textbf{ARC-c} & \textbf{BoolQ} & \textbf{HS} & \textbf{Wino.} \\
\midrule
\rowcolor{white}
Qwen2.5-14B    
& 93.64 & 91.64 & 89.76 & 80.45 & 73.95 \\
\rowcolor{white}
+ \textit{Ours}
& \textcolor{blue}{93.56} & \textcolor{red}{91.98} & \textcolor{red}{89.79} & \textcolor{red}{80.46} & \textcolor{red}{74.27} \\
\midrule
\rowcolor{white}
Mistral-Nemo-Instruct-2407
& 88.34 & 81.14 & 89.76 & 80.40 & 71.19 \\
\rowcolor{white}
+ \textit{Ours}
& \textcolor{red}{88.55} & 81.14 & \textcolor{blue}{87.06} & \textcolor{blue}{80.22} & 71.19 \\
\midrule
\rowcolor{white}
Llama-3.1-8B-Instruct    
& 88.64 & 81.57 & 83.67 & 77.45 & 68.82 \\
\rowcolor{white}
+ \textit{Ours}
& \textcolor{blue}{88.47} & \textcolor{blue}{81.06} & \textcolor{blue}{83.27} & \textcolor{blue}{77.25} & \textcolor{blue}{68.67} \\
\midrule
\rowcolor{white}
Phi-4                   
& 93.52 & 91.98 & 86.18 & 79.31 & 81.45 \\
\rowcolor{white}
+ \textit{Ours}
& \textcolor{red}{93.60} & 91.98 & \textcolor{blue}{86.09} & \textcolor{blue}{79.20} & \textcolor{red}{81.69} \\
\bottomrule
\end{tabular}
\end{adjustbox}
\end{table}

The proposed approach preserves or enhances accuracy in most cases, demonstrating compatibility with diverse model architectures. Where minor performance declines occur, the associated memory and communication efficiency gains provide favorable trade-offs, particularly in resource-constrained edge-cloud deployments. The experimental parameters reflect an optimized balance between quantization aggressiveness and acceptable performance degradation. These results demonstrate that the proposed framework maintains inference accuracy across heterogeneous LLM architectures while consistently reducing server computational requirements, confirming its broad applicability beyond the baseline evaluation models.

\section{Conclusion}

This paper presented an autoregressive-aware split computing framework for deploying large language models (LLMs) on memory- and latency-constrained edge devices. The framework integrated one-point split compression (OPSC) to prevent out-of-memory failures, a two-stage intermediate output compression pipeline (TS+TAB-Q) to preserve accuracy while reducing communication costs, and a unified optimization strategy that jointly determined split placement, quantization settings, and sequence length under strict system constraints.
Extensive experiments across multiple LLMs, hardware platforms, and benchmark tasks demonstrated that the framework consistently outperformed state-of-the-art quantization methods, including Atom, SmoothQuant, and OmniQuant. The framework significantly reduced server-side computation, communication overhead, and end-to-end latency while maintaining or improving accuracy. These results confirmed that the proposed method was effective and scalable for practical edge-cloud LLM deployments.


\bibliographystyle{unsrt}  
\bibliography{references}

\end{document}